\documentclass{article}
\usepackage{ijcai18}

\usepackage{times}
\usepackage{soul}
\usepackage{url}
\usepackage[hidelinks]{hyperref}
\usepackage[utf8]{inputenc}
\usepackage[small]{caption}
\usepackage{graphicx}
\usepackage[normalem]{ulem}
\useunder{\uline}{\ul}{}
\let\ACMmaketitle=\maketitle
\renewcommand{\maketitle}{\begingroup\let\footnote=\thanks \ACMmaketitle\endgroup}
\usepackage{times}
\usepackage{latexsym}
\usepackage{enumitem}

\title{Degree based Classification of Harmful Speech using Twitter Data\footnote{This work was presented at 1st Workshop on Humanizing AI (HAI) at IJCAI'18 in Stockholm, Sweden.}}

\author{
Sanjana Sharma$^1$,
Saksham Agrawal$^2$,
Manish Shrivastava$^3$
\\ 
$^1$IIIT Hyderabad, India\\
$^2$IIIT Hyderabad, India\\
$^3$IIIT Hyderabad, India\\
sanjana.sharma@research.iiit.ac.in,
saksham.agrawal@research.iiit.ac.in,
m.shrivastava@iiit.ac.in
}
\begin{document}

\maketitle

\begin{abstract}
  Harmful speech has various forms and it has been plaguing the social media in different ways. If we need to crackdown different degrees of hate speech and abusive behavior amongst it, the classification needs to be based on complex ramifications which needs to be defined and hold accountable for, other than racist, sexist or against some particular group and community. This paper primarily describes how we created an ontological classification of harmful speech based on degree of hateful intent, and used it to annotate twitter data accordingly. The key contribution of this paper is the new dataset of tweets we created based on ontological classes and degrees of harmful speech found in the text. We also propose supervised classification system for recognizing these respective harmful speech classes in the texts hence.
\end{abstract}

\section{Introduction}

Hate, as a simple standalone word is easily understood by everyone. But as a concept, hate is vast, complex and has multiple themes and extensions. The issue of harmful speech has been widely debated and analyzed by scholars in multiple fields of knowledge. If you’ve been on social media lately, chances are good that you stumbled across something that might be classified as harmful speech online. Perhaps you would have read a tweet that used offensive language to describe its recipient, or maybe you saw a Facebook post that was designed to demean a particular group of people. 

Modern artificial intelligence has proven useful in detecting patterns, whether that be in images for facial recognition or audio for speech regulation. But language is fluid, and as Mark Zuckerberg also recently noted in his testimony\footnote{https://www.nytimes.com/2018/04/10/us/politics/mark-zuckerberg-testimony.html} before the US Congress that harmful speech can be heavily dependent on the context around the hateful words used and intent of the speaker. Some terms found in hate speech are slang, and hence not part of the common vernacular used to train AI. Other pressing issues remain determining different ways of expression of hate and the degree to which it affects people and communities, trying to make a fine line differentiating freedom of speech with hate speech, with making guidelines in defining hate speech \cite{defhatespeech}. Harmful speech has many manifestations, in speeches, prose, literature, real like conversations; and thus it does define it's own certain form in online discourse. It's important for us to understand the amplifications and extensions of harmful speech online, plaguing the social media primarily \cite{understandharmful}. 

Twitter is also actively in an ongoing process to enforce new guidelines related to how it handles hateful conduct and abusive behavior, by users, taking place on its platform. In addition to threatening violence or physical harm, they also want to look for accounts affiliated with certain respective groups that promote violence against citizens to move further in their hateful intentions. Any content that glorifies violence or the perpetrators of a violent act will also be incorporated in violation of Twitter’s new guidelines to combat hate speech. All these new developments in tackling hate speech in online discourse with a spectrum showcasing various degrees and ways (sarcasm, troll, profanity, violent threats etc) manifests a need to studying "Harmful speech online" in detail. This motivated us to develop a classification based on an ontological view of harmful speech, taking inspiration from philosophical and social point of view of hate speech, the intent of speakers involved, affiliation of recipient to an ideology/group or individuality , and deduce them into classes marking some difference in degree of hateful intent. 

\begin{figure*}[h!]
\centering
\includegraphics [width=150mm,scale=0.55] {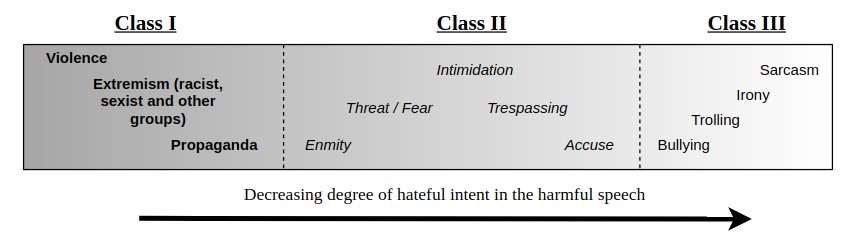}
\caption{Spectrum showing degree based classes of harmful speech}
\end{figure*}

\section{Background and Related Work}

Research on harmful speech has been happening for some time now, but the datasets which are publicly available only classify the text as offensive or not, with a small percentage identifying racist or sexist content. The manual way of filtering out harmful tweets is not scalable, hence it has motivated researchers to identify automated ways. The problem of defining a classification of a tweet and formulating context dependent bad language examples makes the task quite challenging, due to the inherent complexity of the natural language constructs – different forms of hatred, different kinds of targets, different ways of representing the same meaning. 

Combining psychology of hatred with context based classification of hateful speech can prove to reduce the entropy of subjectivity of detection of harmful speech. There are some publicly available datasets that are annotated with degree of hate speech, like \footnote{https://www.kaggle.com/c/jigsaw-toxic-comment-classification-challenge/data} which includes toxic, severe toxic etc as labels. The Kaggle data has around 150k Tweets out of which 16k are toxic, which is around twice the hate speech present in the collected data. But it has classified the degree of hatred into only two distinct classes, severe toxic and toxic, without any granularity like we present in our spectrum classification. To the best of our knowledge, we are the first ones to work on conceptualizing, identifying and classifying ontological classes of harmful speech based on understanding degree of harmful content, intent of the speaker and how it affects people on social media.

\cite{nnaclhlt2016} provided a Hate Speech dataset and the respective annotation procedure in which an initial manual search was conducted in Twitter in order to collect common slurs and terms pertaining to religious, sexual, gender, and ethnic minorities. The main researcher of the article, together with a gender studies student, manually annotated the dataset of 16,918 tweets in categories as racist, sexist and neither of the two. Another article 

\cite{icwww} describes a dataset where messages are classified in the general class “abusive language”, and within the subclasses “hate speech”, “derogatory” and “profanity”. The authors sampled 2,000 comments posted on Yahoo! Finance and News, and noticed that Fleiss’s Kappa value dropped to 0.213 when using the fine-grained three classes (hate speech, derogatory and profanity) as compared to 0.401 for binary classification (only with the class “abusive language”).

The majority of the studies that we found for hate speech were conducted for English. However, some other languages were considered. \cite{arxiv} is an example of a dataset collection and annotation in German, in the specific topic of hate speech against refugees. The results of this study pointed out that hate speech is a vague concept that requires definitions and guidelines in order for having reliable annotations. They also provided solutions for improving this classification task by annotating multiple labels for each tweet, which can be an advantage. Moreover, the authors also note that considering hate speech detection as a regression problem, instead of a binary classification task, can also improve the classifier’s performance.

The most recent paper in hate speech dataset annotation is \cite{DBLP:journals/corr/DavidsonWMW17}. This paper presents data collected using the CrowdFlower platform using hate speech lexicon compiled by Hatebase.org in English. They were instructed that the presence of a particular word, despite being offensive, did not necessarily indicate a tweet is hate speech. They have used the majority decision in CrowdFlower for each tweet to assign a label (Hateful, Offensive or Neither). The intercoder-agreement score provided by CrowdFlower was 92\% and a total percentage of only 5\% of tweets were coded as hate speech by the majority of coders. Consistent with previous work, this study pointed out that certain terms are particularly useful for distinguishing between hate speech and offensive language. Besides, the results also illustrate how hate speech can be used in different ways: it can be directly sent to a person or group of people targeted; it can be espoused to nobody in particular; and it can be used in conversation between people.

\begin{table*}[t!]
\centering
\resizebox{\textwidth}{!}{%
\begin{tabular}{|l|l|}
\hline
Degree & Some examples                                                                                                                                                                                                                                                                                                      \\ \hline
Class I             & \begin{tabular}[c]{@{}l@{}}1. "I fucking hate feminists and they should all be raped and burned to death. \#sluts"\\ 2. "All muslims are faggots  \&  should be slaughtered like pigs." \\ 3. "Now that Trump is president, I'm going to shoot you and all the blacks I can find."\end{tabular}                       \\ \hline
Class II            & \begin{tabular}[c]{@{}l@{}}1.  "You're just an attention whore with no self esteem."\\ 2.  "You are so tan. Ugly dirty bitch. \#LOL"\\ 3. "Shove your opinion up your arse and dance like a monkey."\end{tabular}                                                                                                     \\ \hline
Class III           & \begin{tabular}[c]{@{}l@{}}1. "I hope you all have a great weekend. Except you, Lisa Kudrow. \#CantStandHer"\\ 2. "If you find Benedict Cumberbatch attractive, I'm guessing you'd also quite enjoy staring directly at poop."\\ 3.  "Going to Africa. Hope I don't get AIDS. I am white. \#JustKidding"\end{tabular} \\ \hline
\end{tabular}%
}
\caption{List of examples for categories of Harmful Speech}
\label{my-label}
\end{table*}

\section{Categorization of Hate Speech}

For classifying harmful speech with it's ontological implications, we looked at the philosophical point of view to understand the emotion and verb : hate. Karin Sternberg's theory of hate \cite{{RJSternberg2003}} observed hate as an emotion, a feeling; from an erudite perspective which inspired to form extended nominal categories of hatred. The harmful speech can be molded into a spectrum showing gradient of hate and harm, with some distinguishing characteristics to particular classes and their degrees. The classes have been defined on their decreasing degree of hateful intent from the speaker's perspective, with three classes (Class I, Class II and Class III) showcasing various categories of different types and examples of harmful speech found in social media (like extremism, threatening someone or trolling), as shown in figure 1. As we have devised a spectrum, not a discrete classification, the classes do have an overlapping in their definitions and examples, but in a linear way.

Some of the guidelines used to distinguish the classes are given below. Annotators were asked to keep them in mind along with common sense to classify the respective tweets. Table 1 shows some examples of the three classes thus defined, classified by the annotators subsequently.\\

Class I : 

\begin{itemize}[noitemsep,nolistsep]
  \item Incites violent actions beyond the speech itself.
  \item Is either public or directed at a particular group, mostly with no redeeming purpose. We considered hatred and violent behavior projected to a group to be of more degree than individual accusation and violence. This is an assumption made from the psychological point of view and by seeing various examples that adhere to it. 
  \item The context makes it evident that the speaker wants to intend hurting sentiments of certain isms (extremism) for a violent response to be possible in return.
\end{itemize}

Class II : 

\begin{itemize}[noitemsep,nolistsep]
  \item Cyber banter (accusing, threatening and using aggressive/provocative language for disagreeing  etc.) and verbal dueling constitutes.
  \item The violent characteristic is less than the degree in which Class I operates, which hurts sentiments but not to the degree to invoke a violent response.
  \item Correlates between linguistic violence and non-linguistic/demographic intimidating and trespassing someone in an online space. Can be highly provocative when  addressing an individual rather than some ideology or community/group.
\end{itemize}

Class III : 

\begin{itemize}[noitemsep,nolistsep]
  \item Mildly provocative in nature, mostly given to an individual entity, not necessarily targeting a group or community .
  \item Uses more profane and filthy words not directed or having context from the speaker to the recipient to form a coherent remark. Context mainly revolves around trolling, ironic and sarcastic tone.
  \item Indirect or covert linguistically hurt sentiments, least degree of hateful intent shown in the categories.
\end{itemize}

\section{Corpus Creation and Annotation}
We constructed the corpus using the tweets posted online from Twitter. We had mined tweets with querying for profane slang words and harmful words that we compiled from searching synonyms and various parts of speech extensions of common words that can be used in hateful context. We scraped tweets with hashtags focused on three groups who are often the target of abuse: African Americans (black people), overweight people and women. Some examples of keywords and hashtags handled are : \#IfMyDaughterBroughtHomeABlack, Nigger(s), White Trash, \#IfIWereANazi or the tweets in response to \#MakeAMovieAFatty. Certain hashtags and keywords from recent events surrounding politics, public protests, riots, etc., which have a good propensity for the presence of harmful speech were also used. Certain example of above case can be attributed to the \#GamerGate fiasco, where trolls and haters decided to occupy and corrupt the \#TakeBackTheTech and \#ImagineAFeministInternet hashtags by posting thousands of anti-feminist and misogynistic tweets and memes. We also used resources from Hatebase.org\footnote{https://www.hatebase.org/} to narrow down slangs and hate speech used against the group who are a target of abuse. We retrieved a total of 15,438 tweets from Twitter in json format, which consists of information such as timestamp, URL, text, user, re-tweets, replies, full name, id and likes. A random sampling of the tweets containing respective hashtags and an extensive processing was carried out to remove same tweets (certain reposts of tweets). As a result of manual filtering, a dataset of \textbf{14,906} tweets was created.

\subsection{Annotation}

For the annotation purpose, the annotators selected were well versed with english language as well had a fair amount of experience with witnessing harmful speech online that we deal with on a day-to-day basis on social media. 

\textbf{Harmful Speech or Normal Speech : } A lot of tweets in the dataset collected had occurrence of words which can be used in harmful perspective but didn't evoke any hateful context in the respective tweet at all. Hence, it was required to filter out such normal speech to get a rich dataset of harmful speech only. The initial task in hand was to annotate each tweet with one of the two tags (Harmful Speech or Normal Speech). Harmful speech was detected in \textbf{9064} tweets. Remaining 5842 tweets in the dataset comprised of normal speech, having no context of intent of harm at all, and were of no use for our experiment.

The annotators were provided with a definition along with a detailed guidelines of all the classes and respective categories and examples. Annotators were asked to think about the contextual implications of a tweet, more importantly from the speaker's intent perspective, rather than lexical based judgment of the text, as a syntactic extension of harmful words can not necessarily indicate a tweet inciting hate. It could very well be, plainly stating facts and truth with no intention of hurting sentiments of the recipients. For borderline cases and overlapping examples; sub categories inside the classes (propaganda, enmity, sarcasm etc) their definitions were used, along with measuring relative degree of hateful intent and context, to classify the tweet into a respective class (I, II and III).
Annotation of the corpus was carried out as follows:\newline 

\textbf{Categories of Harmful Speech : } All the 9064 harmful tweets were then manually annotated according to the guidelines mentioned, for classes of harmful speech (Class I, Class II and Class III). The dataset then consisted of \textbf{2138} Class I, \textbf{3924} Class II and \textbf{3002} Class III harmful tweets, after successful annotation. The annotated dataset (consisting of tweet id’s and respective tag of harmful class) with the classification system will be made available online later. All the dataset statistics are shown in Table 2 itself.

\subsection{Inter Annotator Agreement}
In order to validate the quality of annotation, subsequent iterations of annotation was carried out by, in total, 2 human annotators. We calculated the inter-annotator agreement between the two iterations of annotation using Cohen’s Kappa coefficient \cite{doi:10.1177/001316446002000104}. Kappa score was \textbf{0.689} which indicates that the quality of the annotation and presented schema is substantially effective, given how subjective it is to determine the new classification and tuning the degree of harmful content.
\begin{table}[t!]
\begin{center}
\begin{tabular}{|l|rl|}
\hline \bf Categories & \bf No. of tweets & \bf \\ \hline
Class I & 2138 &  \\
Class II & 3924 &  \\
Class III & 3002 & \\
All & \textbf{9064} & \\
\hline
\end{tabular}
\end{center}
\caption{\label{font-table} Twitter harmful speech dataset statistics}
\end{table}

\section{Pre-processing of the tweets}
Given a tweet, we started by applying a light pre-processing procedure based on that reported in \cite{1703.04009}. 
A Twitter-python API\footnote{https://pypi.python.org/pypi/tweet-preprocessor/0.4.0} was used to pre process the tweets as described below :
\begin{enumerate}
  \item \textbf{Removal of URLs and User Names: }All the URLs and links do not contribute towards any kind of sentiment in the text for the tweets. Also, the mentions which are directed to certain users hold no value, hence were also removed.
  \item \textbf{Normalising Hashtags: }Furthermore, we also normalised hashtags into words, so ‘\#killthemuslims’
became ‘kill the muslims’. This is because such hashtags are often used to compose sentences, and we require full words instead, for our method to not miss on the context derived from hashtags too. We used dictionary based look up to split such hashtags, which were made of coherent and correct usage of words.
  \item \textbf{Removal of Special Characters : } All the punctuation marks and special characters (: , ; \& ! ? $\backslash$) in a tweet are also removed.

\end{enumerate}

\section{Experiment and Results}
In this section, we presented our machine learning models which are trained and tested on the respective dataset described in the previous sections. We performed experiments with three different classifiers namely Support Vector Machines with linear function kernel, Naive Bayes method and Random Forest Classifier. For training our system classifier, we have used Scikit-learn \cite{pedregosa2011scikit}. 
In all the experiments, we carried out 10-fold cross validation.
Table 3 describe the accuracy of each model used, in the case of Naive Bayes, Support vector machine and Random forest classifier respectively. The feature set used for SVM and Naive Bayes methods included tf-idf method, and for Random Forest classifier, we used bag of words. Random forest classifier performed the best out of the three and gave a highest accuracy of 76.42\%, while the other two models also gave relevant accuracy with Naive Bayes with 73.42\% and SVM with linear function kernel with 71.71\%.

\begin{table}[t!]
\begin{center}
\begin{tabular}{|l|rl|}
\hline \bf Classifiers & \bf Accuracy & \bf \\ \hline
Naive Bayes & 73.42\% &  \\
Support Vector Machines & 71.71\% &  \\
Random Forest & 76.42\% & \\
\hline
\end{tabular}
\end{center}
\caption{\label{font-table} Accuracy of classification using respective classifiers}
\end{table}

\section{Conclusion and Future Work}

Through this paper, we tried to get the ontological grasp of the hate expression and how it perpetuates it's existence in social media. Harmful speech online is a very subjective domain and has different structures in the social media outreach, from website comments sections to chat sessions in online games. In this paper, we presented an annotated corpus of tweets categorized over various degrees of hate, consisting of tweet ids and the corresponding annotations, in which we tried to give a viable ontological classification model to distinguish harmful speech. We also presented the supervised system used for detection of the class of harmful speech (Class I, Class II and Class III) in the twitter dataset, based on our linear classification skeleton of harmful speech. The corpus consists of 9064 harmful speech tweets annotated with all three classes (degrees) of harmful speech. Best accuracy of 76.42\% was achieved when bag of words approach was used in the feature vector using Random Forest as the classification system. \newline As a part of future work, the supervised methods can be carried out on specific feature set like character n-grams, word n-grams,  punctuation,  negation  words  and  hate lexicon which can give more insight in detailed account of accuracy for each method. \newline The class-based labeling of tweets makes the task in hand very one dimensional, it can be further improved. If the annotation is done by giving scores to each tweet based on degree of hateful intent and other designated characteristics of hate speech in general, the classification problem for automated harmful speech detection and recognition of respective degrees will be considered as a regression model \cite{DBLP:journals/corr/DavidsonWMW17} as compared to a mere classification task to conceptualize linear degree of hateful intent of the text. Various deep learning methods can also be tried and tested on the respective classification to automate the process to some extent.
\newline Our future work includes enlarging and enriching our datasets from social media outlets other than Twitter (example : Reddit) and to work on computationally automatic methods for classifying different forms of harmful speech with subsequent degrees with different methods and this framework a viable scale to distinguish different harmful speech online.

\bibliographystyle{named}
\bibliography{ijcai18}

\end{document}